\documentclass[10pt,twocolumn,letterpaper]{article}

\usepackage[pagenumbers]{cvpr} % To force page numbers, e.g. for an arXiv version

\definecolor{cvprblue}{rgb}{0.21,0.49,0.74}
\usepackage{graphicx}
\usepackage{amssymb}
\usepackage{xcolor}
\usepackage[misc]{ifsym}
\usepackage{array}
\usepackage{multirow}

\usepackage{bbding}
\usepackage{booktabs} % 用于 \toprule, \midrule,
\newcommand{\KVSet}{\mathbf{KV}}
\newcommand{\ModelOuput}{\mathbf{X}_{\theta}}

\newcommand{\KVOutput}{\mathbf{kv}}

\newcommand\blfootnote[1]{%
    \begingroup
    \renewcommand\thefootnote{}\footnote{#1}%
    \addtocounter{footnote}{-1}%
    \endgroup
}
\usepackage[hang]{footmisc}

\usepackage{ulem}
\usepackage{algorithm}
\usepackage{algpseudocode}
\usepackage[pagebackref,breaklinks,colorlinks,allcolors=cvprblue,urlcolor=magenta]{hyperref}
\usepackage[numbers]{natbib}

\newcommand{\coloruline}[2][red]{{%
  \renewcommand{\ULthickness}{1.2pt}%
  \renewcommand{\ULdepth}{2.0pt}%
  \color{#1}\uline{\color{black}#2}\color{black}%
}}
\newcommand{\neguline}[1]{\coloruline[red]{#1}}          % 负向结果

\definecolor{echocolor}{RGB}{30,140,249}
\title{
  {
  Echo\textcolor{echocolor}{\textbf{Torrent}}: Towards Swift, Sustained, and Streaming Multi-Modal Video Generation}
}
\author{
    Rang Meng\textsuperscript{\rm \dag}\,
    Weipeng Wu \,
    Yuming Li\textsuperscript{\rm \ddag}\,
    Chenguang Ma\textsuperscript{\rm \ddag}
    \\
    Ant Group
    \\
}

\begin{document}
\maketitle
% footnote
{
\blfootnote{
    {\rm \dag} Core Contributor \\
    {\rm \ddag} Corresponding author}
}

\begin{abstract}
Recent multi-modal video generation models have achieved high visual quality, but their prohibitive latency and limited temporal stability hinder real-time deployment. Streaming inference exacerbates these issues, leading to pronounced multimodal degradation, such as spatial blurring, temporal drift, and lip desynchronization, which creates an unresolved efficiency–performance trade-off. To this end, we propose \textit{EchoTorrent}, a novel schema with a fourfold design: (1) \textbf{Multi-Teacher Training} fine-tunes a pre-trained model on distinct preference domains to obtain specialized domain experts, which sequentially transfer domain-specific knowledge to a student model; (2) \textbf{Adaptive CFG Calibration (ACC-DMD)}, which calibrates the audio CFG augmentation errors in DMD via a phased spatiotemporal schedule, eliminating redundant CFG computations and enabling single-pass inference per step; (3) \textbf{Hybrid Long Tail Forcing}, which enforces alignment exclusively on tail frames during long-horizon self-rollout training via a causal-bidirectional hybrid architecture, effectively mitigates spatiotemporal degradation in streaming mode while enhancing fidelity to reference frames; and (4) \textbf{VAE Decoder Refiner} through pixel-domain optimization of the VAE decoder to recover high-frequency details while circumventing latent-space ambiguities. Extensive experiments and analyzes demonstrate that EchoTorrent achieves few-pass autoregressive generation with substantially extended temporal consistency, identity preservation, and audio-lip synchronization.
\end{abstract}    
\section{Introduction}
Recent DiT-based audio-driven avatars have achieved remarkable visual fidelity, with smooth motion dynamics and extensive lip-sync consistency across diverse expressions~\cite{wang2025fantasytalking,chen2025hunyuanvideoavatar,yang2024cogvideox,wan2025wan,kong2024hunyuanvideo,wei2025mocha,cui2024hallo3,lin2025omnihuman,chen2025hunyuanvideoavatar,sung2024multitalk,gan2025omniavatar,yang2025infinitetalk,meng2025echomimicv3}. However, they suffer from inflated inference passes and shortened temporal coherence, fundamentally limiting scalability in real-time, long-horizon generation. For instance, state-of-the-art approaches typically require dozens of denoising steps, each involving 2–3 forward passes due to multi-condition classifier-free guidance (CFG), leading to prohibitive latency and computational redundancy. Unfortunately, this substantial computational cost supports only extremely short generations, typically 3 to 5 seconds, highlighting a critical mismatch between model efficiency and the demands of real-world, continuous avatar generation. 

From a first-principles perspective, the most direct path toward real-time, streaming audio-driven avatar generation lies in reducing inference passes and transitioning to an autoregressive formulation. However, in practice, both strategies lead to significant performance degradation, such as (1) visual blurring and texture degradation, (2) long-term identity drift, (3) long-term color drift, and (4) degraded audio-lip synchronization. Moreover, further reducing the number of inference passes within each step, i.e., simplifying multi-condition CFG, undermines the efficacy of existing approaches and amplifies the aforementioned degradations. 

This creates a critical challenge: \textbf{\textit{Is it possible to simultaneously minimize inference passes and multimodal degradations under streaming constraints in audio-driven avatar generation?}}

\noindent\textbf{Multi-Teacher Training.} 
We observe that the teacher model, when used to provide supervision signals for bidirectional-causal distillation, suffers from the following issues: 
(1) a training-inference mismatch under short-sequence chunk configurations; 
(2) a teacher-student post-training gap, wherein the student struggles to acquire enhanced domain-specific abilities (e.g., singing, profile-face speaking, or challenging phoneme lip-sync) after distillation, despite such capabilities being easily accessible to the teacher. 
To address these issues, we propose \textbf{Multi-Teacher Training}, a \textit{SFT-then-RL} framework:  
SFT aligns teacher and student temporal contexts to bridge the training-inference gap;  
RL establishes a specialized teacher ensemble to empower post-distillation specialization on singing, profile views, and rare phonemes.

\noindent\textbf{Adaptive CFG Calibration DMD  (ACC-DMD)} to Further minimize inference passes. Audio CFG exhibits distinct spatial and temporal behavior in contrast to its text counterpart: 
\textup{(i)} Audio CFG prioritizes fine-grained control in facial regions, especially the mouth, with substantial impact on audio-visual synchrony, emotional expression, and even speaker identity; 
\textup{(ii)}~We further observe that different conditional modalities exhibit distinct sensitivity across diffusion timesteps: audio guidance is predominantly effective during the mid-to-early denoising stages. For audio-driven human generation, we leverage the nature of audio CFG to eliminate unnecessary unconditional passes, thereby reducing computational overhead within each denoising step.

To this end, we utilize the schema of ``\textit{decouple, modulate, and remove}'': we first decouple CFG augmentation (CA), the core engine of DMD as highlighted in Z-image-turbo, and then dynamically schedule the CA term across spatiotemporal dimensions. Finally, we remove unnecessary unconditional inference.
Particularly, we adopt teacher-driven CA with a lip-sync-sensitive transfer loss in low-SNR timesteps, while using student-driven CA with an identity-sensitive transfer loss in high-SNR timesteps. By doing so, the spatiotemporal scheduling enables the simultaneous mitigation of global identity degradation and local lip-sync misalignment. Upon a text-CFG-free baseline, our method achieves efficient, high-fidelity audio-driven generation with single-pass few-step inference.

\noindent\textbf{Hybrid Long Tail Forcing}.
Long-term video generation ~\cite{li2025stable,yang2025longlive,huang2025live} has been explored via methods like KV Cache~\cite{huang2025self} and Stream Long Tuning~\cite{yang2025longlive}, but each has notable limitations. KV Cache suffers from a restricted receptive field, failing to align long-term streaming inference effectively. Stream Long Tuning utilizes self-generated rollouts with error propagation as reference frames for distribution alignment on long-horizon rolling extended sequences; however, cumulative errors lead to rigid divergence between the self-rollout and the ground truth, causing residual artifacts and undermining adherence to the reference condition.

On the one hand, we observe that causal attention inevitably introduces degradation in video generation; yet, it is important for efficient KV cache integration in streaming mode. To mitigate the quality drop while retaining streaming compatibility, we propose a causal-bidirectional hybrid attention mechanism, which reduces the adverse effects of pure causal attention while still enabling KV cache for efficient inference. On the other hand, during long-horizon self-rollout, where reference frames suffer from severe degradation, we enforce distillation alignment only on the tail frame of each chunk with the ground truth distribution. This tail-only alignment avoids inconsistency between the entire chunk and the reference frames, thereby preserving temporal coherence and quality. 
To address the spatiotemporal degradation, we induce Hybrid Causal-bidirectional Attention, in which a subset of layers uses causal attention with KV cache to enable streaming, while the remaining layers retain bidirectional attention to preserve global coherence, thus striking a balance between efficiency and fidelity.

\noindent\textbf{VAE Decoder as Refiner}.
Latent alignment mitigates some degradation in streaming inference, but it cannot resolve fine-grained artifacts (identity drift, lip blurring) because VAE compression loses high-frequency details, creating an ambiguous many-to-one mapping. To address this issue directly, we propose extra-latent distillation to align fine-grained distributions beyond the latent space, effectively refining the decoder to suppress accumulated artifacts in the pixel domain. Specifically, we treat the VAE decoder as a refiner. First, we assemble autoregressively generated short clips into extended sequences. These sequences are then aligned at the pixel level with corresponding ground-truth videos using a combination of L1, L2, and  GAN loss, which effectively recovers high-frequency details. The fine-tuned decoder thus suppresses cumulative artifacts and restores sharp, consistent details without adding any computational overhead during inference.

\begin{figure*}[t!]
\begin{center}
\vspace{-0.5em}
\includegraphics[width=1\linewidth]{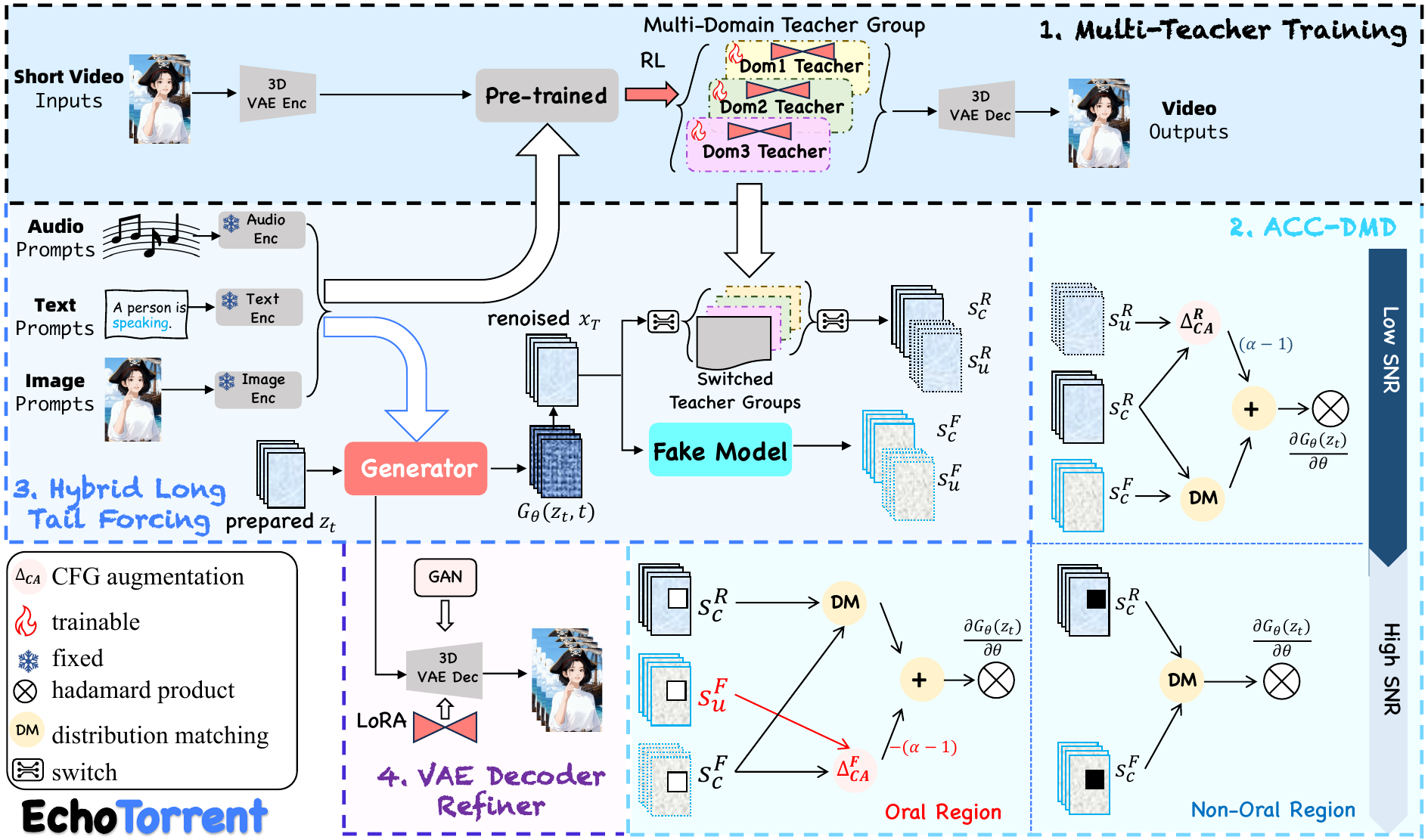}
\end{center}
\vspace{-1.8em}
\caption{The overall training pipeline of EchoTorrent.}
\vspace{-1em}
\label{fig:pipeline}
\end{figure*}
Extensive experiments demonstrate that EchoTorrent effectively suppresses spatial blurring, temporal drift, and lip desynchronization while maintaining high efficiency with few-pass streaming inference.

In summary, our contributions are as follows:
\begin{itemize} 
\item We propose a teacher-student co-optimization paradigm via \textbf{Multi-Teacher Training}, which adapts the teacher model to DM-AR distillation and forms an expert ensemble to transfer preference-diverse knowledge to the distilled student.

\item We introduce the \textbf{ACC-DMD} paradigm, which schedules audio CFG augmentation spatiotemporally to eliminate redundant computations and enable single-pass per-step inference.

\item We present the \textbf{Hybrid Long Tail Forcing} paradigm to enforce alignment exclusively on tail frames during long-horizon self-rollout, mitigating cumulative errors while preserving streaming compatibility. 

\item We also refine the VAE decoder through pixel-domain optimization to recover high-frequency details and suppress fine-grained artifacts without inference overhead. 

\item Our EchoTorrent framework enables streaming, few-pass, long-sequence audio-driven avatar animation, effectively suppressing spatial blurring, temporal inconsistency, and audio-lip desynchronization.
\end{itemize}

\section{Preliminaries}
\subsection{Distribution Matching Distillation}
In the Distribution Matching Distillation (DMD) framework~\cite{yin2024improved,yin2024improved,yin2024onestep,liu2025decoupled}, a student generator, denoted as $G_\theta$, is trained to match the distribution of the pretrained teacher model while reducing the number of inference passes. To this end, DMD
minimizes an approximate KL divergence between the distributions of the student and teacher, whose gradient
is formulated as Eq.~\eqref{equ:dmd_gradient}:
\begin{equation}
\begin{aligned}
\nabla_\theta \mathcal{L}_{\mathrm{DMD}}=\mathbb{E}_{z_t, \tau, x_{\tau}} \left[-(s_{c}^{R}(\mathbf{x}_\tau)-s_{c}^{F}(\mathbf{x}_\tau)) \frac{\partial G_\theta(z_t)}{\partial \theta} \right]
\end{aligned}
\label{equ:dmd_gradient}
\end{equation}
where $z_t$ denotes the prepared latent input with noise level $t$ via backward simulation~\cite{yin2024improved}, and $\mathbf{x}_t$ is obtained by renoising the prediction $G_{\theta}(z_t)$ with noise level $\tau$. The $s_c^{R}$ and $s_c^{F}$ denote the score estimation with conditions $c$ for the teacher and the auxiliary “fake” model that is trained concurrently on the generator’s outputs, respectively. 
$s_{c}^{R}(\mathbf{x}_\tau, t)-s_{c}^{F}(\mathbf{x}_\tau, t)$ quantifies the score discrepancy between teacher and student models. 

In practice, CFG is applied to the teacher model. For the audio-driven avatar generation task, the DMD loss function can be decomposed as the Distribution Matching (DM) term and CFG Augmentation (CA) term, as follows:
\begin{equation}
\nabla_\theta \mathcal{L}_{\mathrm{DMD}}^{\mathrm{CFG}} = 
\mathbb{E} \Biggl[ 
-\biggl(
\Delta_{DM}
+ (\alpha - 1) \Delta_{CA}
\biggr)
\frac{\partial G_\theta(z_t)}{\partial \theta}
\Biggr]
\label{equ:dmd_cfg}
\end{equation}
where $\Delta_{\mathrm{DM}} = s_{c}^{R}(\mathbf{x}_\tau, t) - s_{c}^{F}(\mathbf{x}_\tau, t)$ and $\Delta_{\mathrm{CA}} = s_{c}^{R}(\mathbf{x}_\tau, t) - s_{c}^{R}(\mathbf{x}_\tau)$ denote the distribution matching (DM) and CFG augmentation (CA) terms, respectively. 
The $\Delta_{\mathrm{DM}}$ term strictly enforces the theoretical DMD objective, while $\Delta_{\mathrm{CA}}$ directly injects scaled CFG signals as gradients into the student's output. 
As revealed in~\cite{liu2025decoupled}, the CA term dominates the few-step distillation process, with the DM term acting as a regularizer.

At each training iteration, DMD randomly selects a timestep $\tau \in \{t_1, \dots, t_N\}$ and initializes the reverse process from the corrupted input $\mathbf{x}_{\tau}$, instead of from latent noise $\mathbf{z}$. This strategy effectively simulates mid-generation recovery, enhancing structural consistency across diffusion steps and accelerating convergence.
\subsection{Autoregressive Video Diffusion Models}
% Autoregressive video diffusion models are video generation models that incorporate the iterative reverse diffusion process into the chain-rule factorization of autoregressive models.
In autoregressive video diffusion models, each conditional distribution in the autoregressive chain-rule factorization is modeled by iterative reverse diffusion process~\cite{jin2024pyramidal,teng2025magi,chen2025skyreels,wu2025pack,yin2025slow,deng2024autoregressive,jin2024pyramidal}. Given a video with $N$ frames (or chunks) $\mathbf{x}^{1:N} = \{x^i\}_{i=1}^N$, the joint distribution is factorized via the chain rule into a product of conditional distributions: $p(\mathbf{x}^{1:N}) = \prod_{i=1}^{N} p(x^i \mid \mathbf{x}^{<i})$. Let $p(x^i \mid \mathbf{x}^{<i})$ is the conditional distribution for each frame generation, which is modeled by iteratively denoising process with condition of previously generated frames. The autoregressive denoising process can be formulated as:
\[
p_{\theta}(x^i \mid x^{<i}) = f_{\theta, t_1} \circ f_{\theta, t_2} \circ \ldots \circ f_{\theta, t_N}
\]
where $f_{\theta, t_1}=\Psi(G_{\theta}(x_t^i, t, x^{<i}, c))$, $\Psi$ denotes the forward noising process with timestep $t$, $G_{\theta}$ denotes the ARD model with parameters $\theta$, and $c$ denotes the multi-modal conditions such as text, image and audio. 

\section{Method}
\subsection{Overview}
\noindent\textbf{Pipeline}. In this section, we present EchoTorrent, a hybrid intra- and extra-latent distillation framework for low passes and streaming audio-driven avatar generation, as shown in Fig.~\ref{fig:pipeline}. We propose a two-stage teacher-student co-optimization via \textbf{Multi-teacher Training} (Sec.~\ref{sec:method-initial})
For intra-latent distillation, we propose \textbf{A}daptive \textbf{C}FG \textbf{C}alibration \textbf{D}istribution \textbf{M}atching \textbf{D}istillation (\textbf{ACC-DMD)} via spatiotemporal-aware CFG Augmentation schedule (Sec.~\ref{sec:method-acc-dmd}) to modulate latent distribution alignment effectively. Furthermore, we also improve the streaming long video generation quality via \textbf{Hybrid Long Tail Forcing} (Sec.~\ref{sec:method-long-tail-forcing}). Finally, we employ the 3D causal \textbf{VAE decoder as a refiner} to rectify spatial artifacts such as lips region collapse, as well as identity inconsistency (Sec.~\ref{sec:method-decoder}). Overall, the training procedure and inference procedure can be seen in Algorithm~\ref{alg:algo1} and Algorithm~\ref{alg:algo2} respectively.

\begin{figure}[t]
    \centering
    % \vspace{-0.1in}
\includegraphics[width=\linewidth]{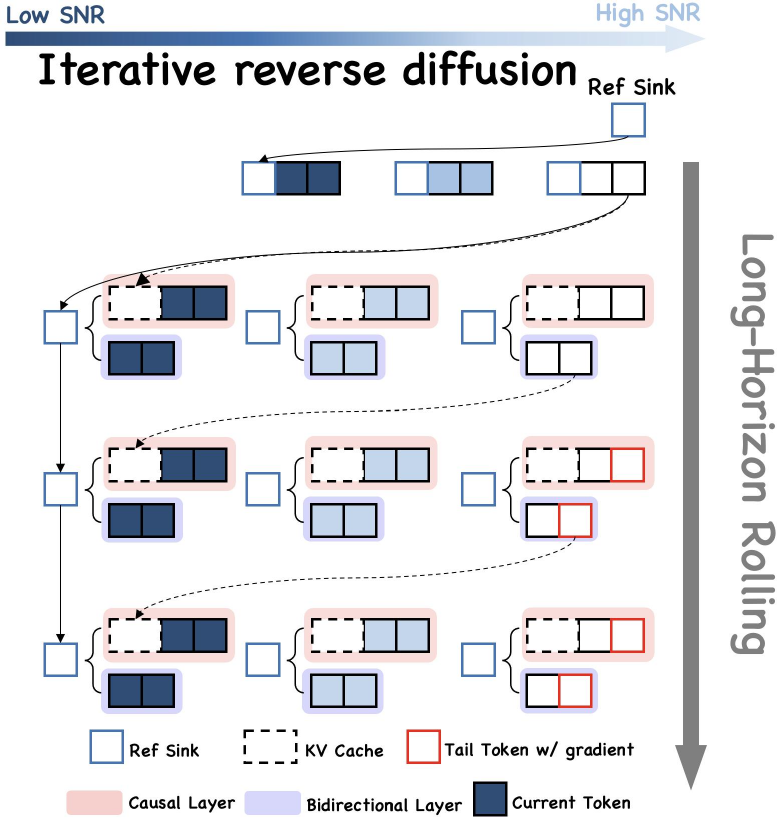}
    \vspace{-0.25in}
    \caption{Hybrid Long Tail Forcing.}
    \vspace{-0.16in}
    \label{fig:Hybrid-Long-Forcing}
\end{figure}
\subsection{Multi-Teacher Training}
\label{sec:method-initial}
Multi-Teacher Training is a \textit{SFT-then-RL} framework:
For SFT, to align the teacher model with streaming inference, we fine-tune it on short chunks of $L_{AR}$ frames, mimicking the actual streaming conditions. This adaptation ensures that the teacher's real scores $s^R$ serve as reliable guidance during distillation. Subsequently, we perform ODE distillation to transfer knowledge from the adapted teacher to the well-initialized student, enabling high-quality, streaming, few-pass avatar generation.

For RL, we fine-tune a pre-trained model on diverse domains via GRPO~\cite{xue2025dancegrpo,guo2025deepseek} to form a teacher ensemble of specialized preference pathways. 
After the initial distillation, this ensemble provides expert instruction on singing lip-sync, profile-view speaking, and rare phoneme lip-sync capabilities, where students typically lag behind teachers.

\subsection{ACC-DMD}
\label{sec:method-acc-dmd}
\textbf{Low-SNR Timestep Distillation.}
Let $\mathcal{T}^{\mathrm{low}}$ denote the set of low-SNR timesteps, corresponding to early denoising stages where the latent retains strong structural priors. For a renoised input $\mathbf{x}_{\tau^{\mathrm{low}}}$ at $\tau^{\mathrm{low}} \in \mathcal{T}^{\mathrm{low}}$, we compute the teacher's real score using audio-guided CFG ($s^R_{\text{cfg}}$) and the fake score using audio conditioning only ($s^F_c$). 
This paradigm strictly adheres to the standard CFG-based DMD framework, where the distillation gradient is scaled by the contrast $\Delta s = s^R_{\text{cfg}} - s^R_0$ between the teacher's conditional ($s^R_{\text{cfg}}$) and unconditional ($s^R_0$) signals, driving $G_\theta$ to learn structurally aligned audio-lip synchrony.

\noindent\textbf{High-SNR Timestep Distillation.}
Let $\mathcal{T}^{\mathrm{high}}$ denote the set of high-SNR timesteps, corresponding to late denoising stages where the latent retains fine-grained details but is vulnerable to identity drift.

\noindent \textbf{Real Score without CFG}. We compute the teacher's real score \textit{without} CFG ($s^R_c$) for two key reasons: 
(i) to prevent the student from inheriting spurious correlations between audio conditions and identity attributes during distillation; 
(ii) to suppress interference from coarse-grained text prompts that may disrupt fine-grained local dynamics.

\noindent \textbf{Fake Score with CFG}. Conversely, we amplify multimodal condition effects in the fake score by applying CFG ($s^F_{\text{cfg}}$), 
thereby inducing a min-max adversarial dynamic between the student and the identity-pure teacher $s^R_c$:
while the CFG-enhanced student gradients increase identity-sensitive interference, the contrast with the purer real score $s^R_c$ (CFG-free) forces the student to resist such biases during distillation. Consequently, the framework mitigates spurious correlations between multimodal conditions and identity consistency during backpropagation.

\noindent \textbf{Oral Region Masked Loss}. To circumvent the potential degradation in lip-sync accuracy of the teacher's real score due to the absence of CFG, we introduce an oral region mask $M_{\text{lip}}: \mathbb{R}^H \times \mathbb{R}^W \to [0,1]^{H \times W}$. This mask enables spatial-aware modulation of the distillation gradient, selectively preserving lip-sync alignment while filtering identity-irrelevant distortions in oral regions. 
The distillation loss for High-SNR timesteps is then formulated as:
\begin{equation}
\small
\begin{aligned}
\nabla_\theta\mathcal{L}_{\text{high-SNR}} = 
\begin{cases} 
\mathbb{E}_{\tau^{\mathrm{high}}} \left[- 
(s^R_c - s^F_{\text{cfg}})\frac{\partial G_\theta(z_t)}{\partial \theta}\right] & \text{if } e \in X_{\text{oral}}, \\
\mathbb{E}_{\tau^{\mathrm{high}}} \left[- 
(s^R_c - s^F_c)\frac{\partial G_\theta(z_t)}{\partial \theta}\right] & \text{if } e \notin X_{\text{oral}}
\end{cases}
\end{aligned}
\end{equation}
where $X_{\text{oral}}$ filters gradients in oral regions of the latent $\mathrm{x}_\tau$ by masking the teacher-student discrepancy when CFG is disabled for the teacher, preventing suboptimal lip dynamics from transferring to the student while allowing identity constraints to operate globally.

\subsection{Hybrid Long Tail Forcing}
\label{sec:method-long-tail-forcing}
\noindent \textbf{Long-horizon Self-rollout}.
Given a long video $V_N$ where $N > w$ ($w$ is short window video length), we first apply ACC DMD over the initial window $W_0$. Subsequently, the KV cache of the self-generated tail blocks is stored and used as the starting context for the next window $W_{i+1}$. This rolling generation proceeds across $V_N$ until reaching the maximum length, and then a new batch is resampled.

\noindent\textbf{Tail-only Forcing}. As the rolling index $i$ increases, accumulated errors cause growing divergence between the synthesized clip $W_i$ and the distribution of teacher model. Since our framework uses the first frame as the reference for consistency within each window, naively enforcing alignment on the entire window would compromise this reference-based mechanism, leading to degraded reference fidelity and reduced diversity in generation. To this end, we enforce ACC DMD only on the last frame of each window $W_i$ with the corresponding pretrained distribution. By doing so,  most frames within each window preserve reference-based consistency, whereas the inter-window cumulative drift, pivotal for streaming video coherence,is involved and rectified during training.

\noindent\textbf{Causal-Noncausal Hybrid Attention.} We replace the causal-attention layers in the initial autoregressive student model $G_{\theta}$ with bidirectional ones for selected layers, creating a hybrid architecture that balances streaming efficiency (via causal-attention) and quality (via Noncausal-attention). This hybrid design mitigates degradation while still enabling long-horizon rolling KV cache for training-test alignment. 
\begin{algorithm}[H]
\caption{EchoTorrent Forcing Training}
% with KV Recaching
\begin{algorithmic}[1]
\Require Hybrid generator $G_\theta$, Condition set $C$ (including text and audio)
\Require Video length $N$, Per clip length $w$, reference sink cache $\mathbb{S}$, Long Sequence threshold $\hat{N}$
\While{not converged}
    \State Initialize KV cache $\mathbb{KV} \gets []$
    \State Initialize current video length $l \gets 0$
    \State Sample switch index $i$
    \Statex \hspace{\algorithmicindent}$i \in \{1,2,\dots,\lfloor L/w\rfloor-1\}$
    \State $s \gets i \cdot w$
    
    \If{$l \ge w$}
      \State $\mathbb{KV} \gets []$;\quad $l \gets 0$
      \State Resample $i$
    \EndIf
    
    \If{$l = s$}
      \State $\mathbb{KV} \gets \texttt{recache}(G_\theta,C,\mathbb{KV},\mathbb{S})$ 

    \EndIf
    
    \State $W_{i}^{1:w} \gets \texttt{generate\_next\_clip}(G_\theta,C,\mathbb{KV},\mathbb{S})$ 
    \If{$l \ge \hat{N}$}
    \State $\mathcal{L} \gets \texttt{ACC-DMD}(G_\theta,\ W_{i}^{tail},\mathbb{KV},\mathbb{S})$
    \Else
        \State $\mathcal{L} \gets \texttt{ACC-DMD}(G_\theta,\ W_{i}^{1:w},\mathbb{KV},\mathbb{S})$
    \EndIf
    \State $\mathcal{L}\texttt{.backward}()$
    \State update generator parameter $\theta$
\EndWhile
\end{algorithmic}
    \label{alg:algo1}
  \end{algorithm}

\noindent\textbf{Bidirectional Reference Cache.} 
Notably, our Non-causal attention is not fully bidirectional; instead, we augment both causal and non-causal attention layers with fixed reference frame sinks (cached features). These sink tokens have unidirectional connections to the original tokens with bidirectional attention, providing stable reference without leaking future information.
% \begin{algorithm}[t]
% \caption{EchoTorrent Inference}
% \small
% \begin{algorithmic}[1]
%   \Require Per-timestep KV caches, each with size $L$
%   \Require Timesteps $\{t_1, \dots, t_T\}$
%   \Require Number of generated frames $M$
%   \Require Conditions of $N$ frames $C_{1:N}$ (including audio,text)
% \Require Ref image $R$ 
%   \Require Flow-Matching Model $v_\theta^\text{KV}$ (extra returns KV embeddings)
%   \Require Rolling RoPE transform $\Phi(\cdot)$ 
%   \Require VAE Decoder $\text{VAE}(\cdot)$ 
%   \State Initialize model output $\ModelOuput \gets []$
%   \State Initialize KV caches $\{\KVSet_1, \dots, \KVSet_T\} \gets []$
%   \State Initialize Rolling Sink Frame $ \SinkFrame \gets R$
%   \State Initialize $dt \gets -1/T$
%   \For{$i = 1, \dots, M$}
%     \State Initialize $x^i \sim \mathcal{N}(0, I)$
%     \For{$j = T, \dots, 1$}
%       \State Set $\hat{v}^i_{j},\KVOutput^i_j \gets v_\theta^\text{KV}(x^i; t_j, \KVSet_j,C_i,\Phi(\SinkFrame))$ \Comment{RoPE Update}
%       \State Set $x^i \gets x^i + \hat{v}^i_{j} dt$
%     \If{$|\KVSet_j| = L$}
%       \State $\KVSet_j.\mathrm{pop}(0)$ 
%     \EndIf
%     \State $\KVSet_j{\texttt{.append}}(\KVOutput^i_j)$
%     \EndFor
%     \State $\ModelOuput{\texttt{.append}}(  \text{VAE}(x^i))$
%     \If{$i = 1$}
%         \State $\SinkFrame \gets  x^i$  \Comment{AAS Update}
%     \EndIf
%   \EndFor
%   \State \Return $\ModelOuput$
% \end{algorithmic}
% \label{alg:inference1}
% \end{algorithm}
% % \end{minipage}
\begin{algorithm}[H]
    \caption{EchoTorrent Long-Horizon Inference}
    \small
    \begin{algorithmic}[1]
      \Require Per-timestep KV caches, each with size $k$
      \Require Timesteps $\{t_1, \dots, t_T\}$
      \Require Number of generated frames $L$
      \Require Conditions $C_{1:N}$ (including audio,text)
      \Require Reference image $I$ 
      \Require Hybrid generator $G_\theta^\text{KV}$ (extra returns $\mathbb{KV}$ embeddings)
      \Require Rolling RoPE transform $\Phi(\cdot)$ 
      \Require Refined VAE Decoder $\text{VAE}(\cdot)$ 
      \State Initialize model output $\ModelOuput \gets []$
      \State Initialize $\mathbb{KV}$ caches $\{\mathbb{KV}_1, \dots, \mathbb{KV}_T\} \gets []$
      \State Initialize Rolling Sink Frame $ \mathbb{S} \gets I$
      % \State Initialize $dt \gets -1/T$
      \For{$i = 1, \dots, M$}
        \State Initialize $x^i \sim \mathcal{N}(0, I)$
        \For{$j = T, \dots, 1$}
          \State Set $x^i,\KVOutput^i_j \gets G_\theta^\text{KV}(x^i; t_j, \KVSet_j,C_i,\Phi(\mathbb{S}))$
          % \State Set $x^i \gets x^i + \hat{v}^i_{j} dt$
        \If{$|\mathbb{KV}_j| = L$}
          \State $\mathbb{KV}_j.\mathrm{pop}(0)$ 
        \EndIf
        \State $\mathbb{KV}_j{\texttt{.append}}(\KVOutput^i_j)$
        \EndFor
        \State $\ModelOuput{\texttt{.append}}(  \text{VAE}(x^i))$
        \If{$i = 1$}
            \State $\mathbb{S} \gets  x^i$  
        \EndIf
      \EndFor
      \State \Return $\ModelOuput$
    \end{algorithmic}
    \label{alg:algo2}
  \end{algorithm}

\vspace{-0.1in}
\subsection{VAE Decoder Refiner}
\label{sec:method-decoder}
Intra-distillation suffers severe detail degradation in long-horizon avatar animation, particularly under \textit{few-pass} and \textit{streaming} settings: 
1) Few-pass timestep skipping favors low-frequency structural regions but breaks fine-grained details; 
2) Streaming accumulates latent-space errors, rendering direct detail recovery infeasible due to the fragility of compressed features. 
To recover high-frequency details that latent-space alignment fails to capture, we perform distillation directly in pixel space. We generate consecutive segments $\mathbf{W}$ from the frozen student generator $G_{\theta}$, concatenate them to a fixed length $L$ noted as $\mathbf{Z}^R$, and decode them via the VAE decoder $\mathcal{D}$. The resulting pixel sequences are then optimized against the corresponding ground-truth frames $\mathbf{Z}_{\text{gt}}$ using a combination of LPIPS loss $\mathcal{L}_{\text{LPIPS}}$ and adversarial loss $\mathcal{L}_{\text{adv}}$:
\begin{align}
&\mathcal{L}_{\text{extra}} = \mathcal{L}_{\text{LPIPS}} + \mathcal{L}_{\text{adv}} \\
&\mathcal{L}_{\text{adv}} = \mathbb{E}_{\mathbf{W}} \left[ \log D(\mathbf{Y}^{R}) + \log(1 - \mathcal{C}(D(Z_{L})) \right]
\end{align}
where $\mathbf{Y}^{R}$ denotes the real frames, $\mathcal{C}$ denotes the concatenation operation, and $D_{\text{VAE}}$ is the discriminator attached to the decoder. This pixel-level refinement suppresses cumulative artifacts and enhances visual fidelity without adding inference cost. During this stage, only $\mathcal{D}_{\text{VAE}}$ and $D_{\text{VAE}}$ are updated while $G_{\theta}$ remains frozen.
\begin{table*}[t!]
\centering
\caption{Quantitative Comparisons with SOTAs.}
\vspace{-0.07in}
\resizebox{\textwidth}{!}{
\begin{tabular}{ll|cc|ccccccccc}
\hline\hline
duration  & Method&\#Params& FPS  & Sync-C$\uparrow$ & Sync-D$\downarrow$ & FID$\downarrow$ & FVD$\downarrow$ & IQA$\uparrow$ & ASE$\uparrow$ & ID$\uparrow$ & HA$\uparrow$ & BC$\uparrow$  \\
\hline
\multirow{7}{*}{Short sequence} 
& EchoMimicV3-Flash~\cite{meng2025echomimicv3} &1.3B  &  2.12   &  6.13   &  9.09   & 42.76 &496.88    &4.96   & 3.75  &0.89 &0.88&0.92 \\
& HunyuanAvatar~\cite{chen2025hunyuanvideo}    &14B  &    1.31 &6.12  & 9.11 &42.54 &676.28 &4.96 &3.67&0.89  &0.85&0.89 \\
& FantasyTalk~\cite{wang2025fantasytalking}    &14B  &    1.31 &4.05  &11.01 &45.03 &603.95 &4.85 &3.48&0.86 &0.80&0.87 \\
& InfiniteTalk~\cite{yang2025infinitetalk}     &14B  &    1.31 &6.97 &8.65 &38.56 & 418.43 &5.07  &3.97&0.91 &0.90&0.96 \\
& SoulX-FlashTalk~\cite{shen2025soulx}& 14B   &  8.21   &6.65     &8.69 &38.89  &418.67  & 5.02   &3.86  &0.92 &0.89&0.95 \\
\hline
& \textsc{EchoTorrent} &14B  &10.50  &6.77 &8.52     &38.85     &415.76     &5.09 &3.98   &0.92 &0.91&0.96 \\
\hline\hline
\multirow{5}{*}{Long Sequence}  
& InfiniteTalk    &14B  &1.31  &6.88  &8.79  &40.08  &431.09 &4.97 &3.90&0.80 &0.88&0.90  \\
&LiveAvatar~\cite{huang2025live}&14B  &5.35   &5.80   &9.58  &42.87  &500.09 &4.90 &3.68 &0.76  &0.84&0.88 \\
& SoulX-FlashTalk&14B   &8.21   &6.62  &8.70 &38.89  &420.65  & 4.99  &3.90  &0.81  &0.87&0.91 \\
\hline
&\textsc{EchoTorrent}  &14B   &10.50 &6.71   &8.60 &39.01  &420.17  &5.01  &3.92  &0.87   &0.89&0.93 \\
\hline\hline
\end{tabular}
}
\vspace{-0.05in}
\label{tab:1}
% \vspace{-0.15in}
\end{table*}

\begin{figure*}[t]
\centering
% \vspace{-0.1in}
\includegraphics[width=1.0\linewidth]{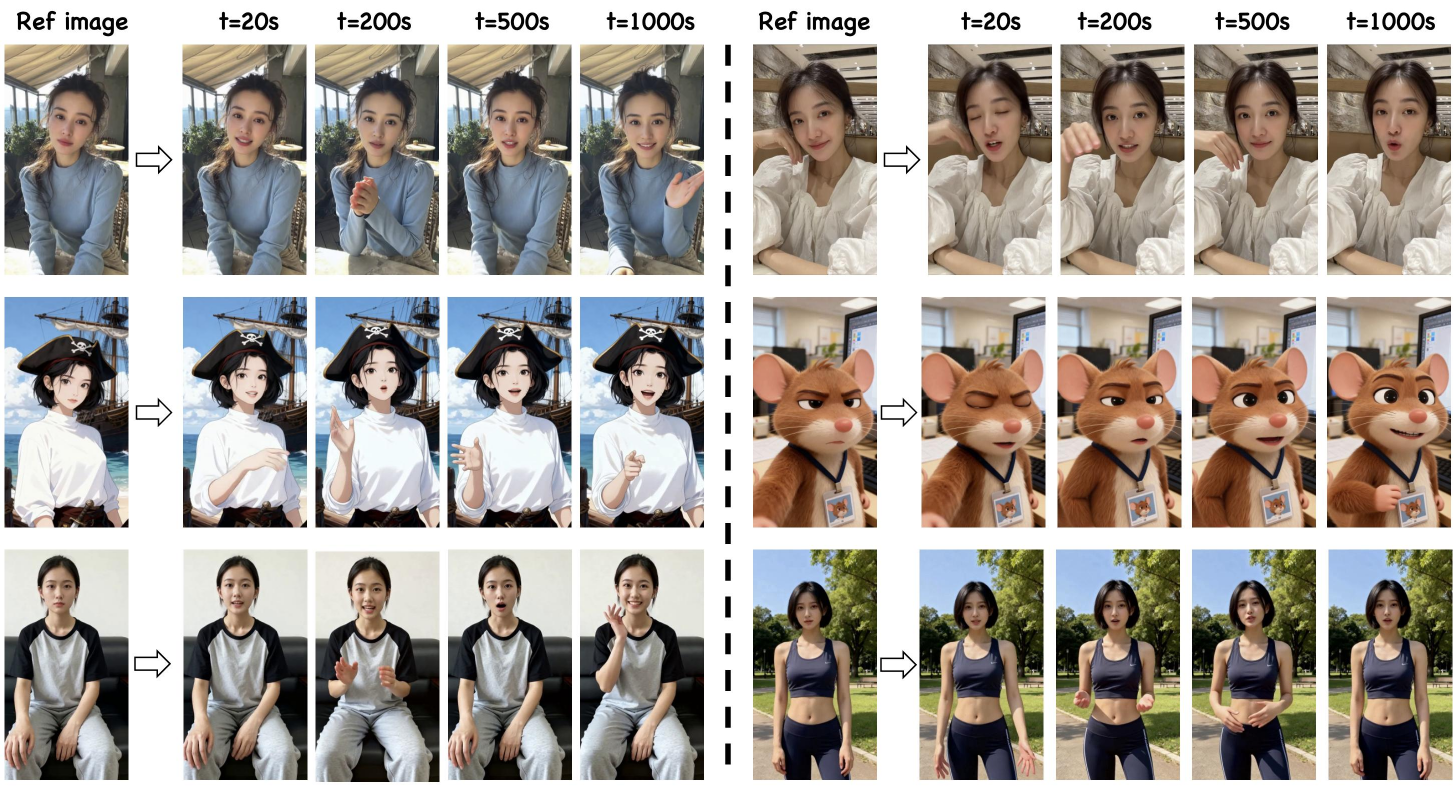}
\vspace{-0.25in}
\caption{Qualitative Results for Long-Horizon Robustness. Visualizations for different duration ranging from 20 seconds to 1000 seconds.}
\vspace{-0.1in}
\label{fig:long}
\end{figure*}

\section{Experiments}
\subsection{Experimental Settings}
\noindent\textbf{Implementation.}
The teacher and student models share this baseline architecture from InfiniteTalk~\cite{yang2025infinitetalk}. After Multi-Teacher Training, we adapt the teacher and student models from LightX2V~\footnote{https://huggingface.co/Kijai/WanVideo\_comfy/tree/main/Lightx2v}, a few-step image-to-video diffusion model. The generated videos have a spatial resolution of $400\times720$ and a total duration of 3 latent frames. For ACC-DMD, we set the audio CFG scale to 1.8. In autoregressive streaming, each generated chunk consists of 4 latent frames, and we maintain a KV cache of the most recent 12 frames. To balance causality and quality, we convert causal attention layers to bidirectional attention in a staggered pattern (every other layer), resulting in a hybrid attention mechanism. Experiments are conducted on 64 PPUE GPUs, with 5K iterations for teachers training, 10K iterations for ACC-DMD, 5K iterations for Hybrid Long Tail Forcing, and 2K iterations for VAE Decoder refining, respectively. The batch size is set to 1 for each GPU. The learning rate is set to 2e-6 for the student and 4e-7 for the fake score.

\noindent\textbf{Inference Acceleration.} 
We benchmark throughput on 8 A100 (\textbf{\textit{our maximum accessible resource configuration}}) GPUs, using a 4-NFE sampling schedule. We implement sequence parallelism using Ulysses in xDiT to reduce communication overhead. Additionally, the VAE decoder is parallelized along the temporal axis, enabling the efficient generation of long video sequences.

\noindent\textbf{Datasets.}
For Streaming Enhancement Initialization, We train our model on a combination of the EchoMimicV3 dataset, the HDTF dataset~\cite{zhang2021flow}, and additional self-collected data, a large-scale audio-visual corpus containing numerous video clips of human subjects with synchronized audio. After filtering, we use 300K samples for training.

\noindent\textbf{Metrics.} 
To comprehensively evaluate our model's performance, we employ the following metrics:
1) Image Quality: Fréchet Inception Distance (FID)~\cite{deng2019accurate} is used to assess the fidelity and visual quality of generated images.
2) Video Quality: Fréchet Video Distance (FVD)~\cite{unterthiner2018towards} measures temporal coherence and overall video quality.
3) Perceptual and Aesthetic Analysis: We analyze perceptual quality (IQA) and aesthetic appeal (ASE) of the generated content.
4) Audio-Visual Alignment: For lip-sync tasks, we use Sync-C and Sync-D metrics~\cite{prajwal2020lip} to evaluate synchronization accuracy.
5) Consistency Metrics: We adopt Vbench2.0~\cite{zheng2025vbench} metrics, including Identity Consistency (ID), Human Anatomy (HA), and Background Consistency (BC).
For quantitative evaluation, we randomly select 400 videos generated by EchoTorrent. Note that our HA metric emphasizes booth and hand details, aligning with human perceptual focus on audio-driven avatar animation.
\subsection{Comparison with SOTA Methods}
We evaluate our framework against leading half-body digital human generation techniques: EchoMimicV3-Flash\cite{meng2025echomimicv3}, InfiniteTalk\cite{sung2024multitalk}, LiveAvatar, and SoulX-FlashTalk. Quantitative results in Table\ref{tab:1} demonstrate that our method is on par with or exceeds the state of the art in visual quality, synchronization, and temporal efficiency. 
Notably, EchoTorrent sets a new benchmark in audio-lips sync and video quality for short-duration generation. In visual quality assessment, EchoTorrent achieves an IQA score of 5.09, surpassing SoulX-FlashTalk (5.02). On Sync-C, a motion-aware metric for lip-sync precision, our model reaches 6.77, exceeding InfiteTalk’s 6.67 and highlighting improved dynamic correspondence.
Efficiency is a critical factor in real-time applications. EchoTorrent operates at 10.50FPS on a 14B-parameter architecture with 8 A100 GPUs.
For long-form generation, sustained synchronization is crucial. EchoTorrent achieves a Sync-C of 6.71 and Sync-D of 8.60, outperforming InfiniteTalk, LiveAvatar and SoulX-FlashTalk, indicating stronger preservation of audio-visual alignment over extended sequences. Notably, the system maintains a consistent 10.50FPS throughput even during prolonged inference, confirming its scalability and temporal robustness. We attribute this stability to the proposed training schema, which mitigates error accumulation and temporal drift commonly observed in hybrid-attention streaming models.
These results collectively demonstrate that EchoTorrent not only enables expressive, full-body audio-driven animation but also achieves state-of-the-art performance across quality, synchronization, and efficiency metrics, without compromising real-time capabilities.

\subsection{Ablation Studies}
\noindent\textbf{Ablation on Streaming Enhancement Initialization (SEI).}
We validate the effectiveness of our SEI paradigm for EchoTorrent. As shown in rows $3$ of Table \ref{tab:ab}, we observe that Teacher ST has a significant impact on the visual quality metric (IQA), audio-lips sync and the human motion metric (HA) because it adapts the teacher model to accommodate shorter sequence generation, thereby providing a more accurate real score for the streaming model. Furthermore, the row $4$ indicates that AR initialization for generator alleviates the degradation in almost all spatiotemporal metrics such as IQA, Sync-C, ID, HA, and BC.

\begin{table}[t!]
\centering
\vspace{-0.08in}
\caption{Ablation study for EchoTorrent in Long-Horizon Generation.}
\vspace{-0.08in}
\resizebox{.48\textwidth}{!}{
\begin{tabular}{l|cccccccc}
\hline
Components & Sync-C$\uparrow$ & IQA$\uparrow$ & ID$\uparrow$ & HA$\uparrow$ & BC$\uparrow$ \\
\hline\hline
EchoTorrent &6.71 &5.01 & 0.91 &0.89 &0.93 \\
\hline\hline
w/o SEI  &\neguline{6.65}   &\neguline{4.90} &\neguline{0.86} &\neguline{0.84} &0.92  \\
\hline\hline
w/ original DMD &\neguline{6.45} &5.01 &\neguline{0.87} &0.89 &0.93 \\
\hline\hline
Full Causal   &6.71 &\neguline{4.70} &0.91 &\neguline{0.81}  &\neguline{0.89}  \\
Full Bidirectional   &6.71 &4.99  &\neguline{0.86} &\neguline{0.87} &0.93  \\
w/o Reference Sink  &6.71&4.99 &\neguline{0.85} &0.89  &0.93  \\
All Forcing         &\neguline{6.60}   &\neguline{4.90}   &\neguline{0.86}  &\neguline{0.86} &\neguline{0.90}  \\
\hline\hline
w/o Decoder FT   &6.71   &\neguline{4.61}   &0.90  &\neguline{0.84} &0.92  \\
\hline
\end{tabular}
}
\vspace{-0.25in}
\label{tab:ab}
\end{table}

\noindent\textbf{Ablation on ACC-DMD.} Results of the rows $5$ in Table \ref{tab:ab} demonstrate that the approach without ACC-DMD achieves both suboptimal lip synchronization and identity preservation. This result indicates that audio injection also weakens the temporal coherence of the reference image.

\noindent\textbf{Ablation on Hybrid Long Tail Forcing.} We evaluate components of our Hybrid Long Tail Forcing framework in three aspects: 1) Causal-Bidirectional Hybrid Attention, 2) Long Tail Forcing training, and 3) reference sink. As shown in Rows 6–9 of Table \ref{tab:ab}, full attention architectures (causal or non-causal) provide inadequate suppression of degradation (detail, motion and identity) compared to our Hybrid design. This stems from dual reasons: i) full bidirectional attention fails to fully leverage the informational gains from previous frames via KV cache to mitigate accumulated errors; ii) while full causal attention disrupts pretrained temporal correlations, amplifying errors within a single chunk. Results in row $8$ showcase that the reference sink can boost the identity preservation metric without compromising visual quality, lip-sync, and motion anatomy metrics.

\noindent\textbf{Ablation on the VAE Decoder Refiner} (Table \ref{tab:ab}, last row) confirms its capacity to enhance perceptual quality while maintaining parity in other metrics.

\noindent\textbf{Ablation on Long-Horizon Robustness.} We further evaluate the long-horizon robustness of EchoTorrent. As shown in Figure~\ref{fig:long}, EchoTorrent achieves temporal coherence of up to 1,000 s. Critically, perceptual quality, subject deformation, background naturalness, and identity consistency are maintained with extended generation durations.
\vspace{-0.1in}
\section{Related Work}
\subsection{Multimodal Video Generation}
\textbf{Video Generation.} 
The field of video generation has evolved from early 2D CNN architectures~\cite{} augmented with handcrafted motion priors to recent DiT diffusion frameworks~\cite{HaCohen2024LTXVideo,wang2023modelscope,ho2022video,yang2024cogvideox,consisid,phantom,vace,animateanyone,followyourpose,controlavideo,controlvideo,text2videozero,videop2p,lin2024open,openaisora2024,CDT,xu2024easyanimate,xu2024easyanimate,bao2024vidu,zhipu2024qingying}. While spatiotemporal convolutions enabled models such as SVD~\cite{blattmann2023stable} to generate short clips with high spatial fidelity, transformer-based diffusion models have significantly advanced temporal coherence and motion smoothness by replacing hard-coded inductive biases with data-driven self-attention mechanisms.
Recent progress in text-to-video generation is largely driven by large-scale pretraining~\cite{minimax2024hailuo,tongyi2024wanxiang,pika2024pika,zheng2024open,step_video,wang2023videolcm,wang2024unianimate,liu2024timestep,unterthiner2018towards,wu2024improved}. CogVideoX~\cite{yang2024cogvideox} achieves fine-grained semantic alignment through hierarchical token fusion; HunyuanVideo~\cite{kong2024hunyuanvideo} introduces a dual-stream DiT architecture to enhance conditional control robustness; and Wan~\cite{wan2025wan} demonstrates that scaling model capacity alone can yield state-of-the-art visual fidelity, setting new benchmarks for general-purpose video synthesis.\\
\textbf{Audio-Driven Avatar Animation.} 
Audio-driven avatar generation aims to synthesize expressive and temporally coherent human performances from speech signals~\cite{gao2025wan,lin2025omnihuman,ma2025controllable,tu2025stableavatar,jiang2025omnihuman,cui2025high,team2025klingavatar,wang2025interacthuman,sun2025streamavatar,huang2025hunyuanvideo,li2025infinityhuman,liang2025alignhuman,du2025rap,zhang2025soul,wang2025rest,li2025joyavatar,zhong2025anytalker,ma2025playmate2,liang2026integrating}. While early approaches primarily focused on talking heads~\cite{zhu2024champ,xu2024vasa}, recent advances have extended to full-body animation with improved realism and control. Representative methods include EMO~\cite{tian2024emo}, which employs temporal encoding for dynamic expression modeling; AniPortrait~\cite{wei2024aniportrait}, which leverages 3D-to-2D pose guidance for natural body motion; Vlogger~\cite{zhuang2024vlogger}, which adopts diffusion-based rendering for high-fidelity synthesis; EchoMimic~\cite{chen2024echomimic} which enables multimodal conditioning for diverse driving inputs; and EchoMimicV2~\cite{meng2024echomimicv2}, which first extends face generation to upper-body human generation. CyberHost~\cite{lin2024cyberhost} further supports rich user controls such as hand gestures and body trajectories, while OmniHuman~\cite{lin2025omnihuman} improves long-term stability via progressive dropout. For multi-character scenarios, MultiTalk~\cite{sung2024multitalk} and HunyuanVideo-Avatar~\cite{chen2025hunyuanvideoavatar} introduce interaction-aware modeling, and OmniAvatar~\cite{gan2025omniavatar} proposes efficient audio injection into transformer blocks for low-latency inference.

\subsection{Real-time Multi-modal Video Generation}
Real-time video generation has witnessed rapid advancements recently~\cite{chen2024diffusion,song2025history,teng2025magi,guo2026efficient,chen2026context,lu2025reward}. For instance, CausViD~\cite{yin2025causvid} proposes to leverage DMD to transform bidirectional attention into unidirectional attention for streaming video generation. Self-Forcing~\cite{huang2025self} introduces the use of KV cache during training to expand the receptive field and better align training with inference. Self-Forcing++~\cite{cui2025self} further extends this framework with re-caching strategies. LongLive~\cite{yang2025longlive} presents a "train-long-test-long" paradigm, incorporating long-sequence training with degradation noise and integrating frame sink mechanisms to improve identity consistency. 

On the other hand, real-time audio-driven avatar animation has also seen rapid progress. Live Avatar~\cite{huang2025live} proposes an algorithm-system co-design framework that enables efficient streaming inference for avatar animation. Knot Forcing~\cite{xiao2025knot} dynamically updates the temporal position of reference frames and introduces temporal knots as chunk-wise bridges to enhance temporal coherence across autoregressive steps. LiveTalk~\cite{chern2025livetalk} trains the ODE initialization to convergence, providing a robust initialization for causal attention to better mitigate error accumulation. SoulX-FlashTalk~\cite{shen2025soulx} leverages intra-chunk bidirectional attention within an autoregressive diffusion framework, enabling richer context modeling while maintaining causal inference constraints.

\section{Conclusion}
In this work, we propose EchoTorrent, a 4-NFE streaming framework for high-quality, long-horizon, multi-modal video generation. We introduce a teacher-student-Decoder collaborative post-training paradigm, design a novel hybrid causal-bidirectional architecture, and propose two novel distillation approaches (i.e., ACC-DMD and Long Tail Forcing) to address degradation under joint constraints of few NFE, streaming mode, and long-sequences. Comprehensive experiments and ablation studies demonstrate that EchoTorrent exceeds current state-of-the-art methods in both quantitative metrics and qualitative evaluations, achieving superior performance in real-time, high-quality, and long-horizon multi-modal human video generation.
{
    \small
    \bibliographystyle{IEEEtran}
    \bibliography{main}
}

\end{document}